\documentclass{scrartcl}
\usepackage[T1]{fontenc}
\usepackage[utf8]{inputenc}
\usepackage{amsmath, amsthm, amssymb}
\usepackage{dsfont}
\usepackage{graphicx}
\usepackage[format=plain]{caption}
\usepackage{capt-of}
\usepackage{subcaption}
\usepackage{bm}
\usepackage{tikz}
\usetikzlibrary{arrows,shapes,trees,calc,through}
\usepackage{csquotes}

\renewcommand{\phi}{\varphi}

\numberwithin{equation}{section}

\theoremstyle{definition} 
\theoremstyle{definition} 
\theoremstyle{definition}

\begin{document}

\title{Capturing and incorporating expert knowledge into machine learning models for quality prediction in manufacturing}

\author{Patrick Link$^1$, Miltiadis Poursanidis$^2$, Jochen Schmid$^2$,  \\Rebekka Zache$^1$, Martin von Kurnatowski$^2$, Uwe Teicher$^1$, \\Steffen Ihlenfeldt$^{1,3}$ \\  
\small $^1$ Fraunhofer Institute for Machine Tools and Forming Technology (IWU), 09126~Chemnitz,\\ \small Germany\\
\small $^2$ Fraunhofer Institute for Industrial Mathematics (ITWM), 67663 Kaiserslautern, Germany\\ 
\small $^3$ Institute of Mechatronic Engineering, Technische Universit\"at Dresden, 01062 Dresden, \\ \small Germany\\
\small patrick.link@iwu.fraunhofer.de}  

\date{}

\maketitle

\begin{abstract}
\small{ \noindent 
Increasing digitalization enables the use of machine learning methods for analyzing and optimizing manufacturing processes. A main application of machine learning is the construction of quality prediction models, which can be used, among other things, for documentation purposes, as assistance systems for process operators, or for adaptive process control. The quality of such machine learning models typically strongly depends on the amount and the quality of data used for training. In manufacturing, the size of available datasets before start of production is often limited. In contrast to data, expert knowledge commonly is available in manufacturing. Therefore, this study introduces a general methodology for building quality prediction models with machine learning methods on small datasets by integrating shape expert knowledge, that is, prior knowledge about the shape of the input-output relationship to be learned. The proposed methodology is applied to a brushing process with $125$ data points for predicting the surface roughness as a function of five process variables. As opposed to conventional machine learning methods for small datasets, the proposed methodology produces prediction models that strictly comply with all the expert knowledge specified by the involved process specialists. In particular, the direct involvement of process experts in the training of the models leads to a very clear interpretation and, by extension, to a high acceptance of the models. Another merit of the proposed methodology is that, in contrast to most conventional machine learning methods, it involves no time-consuming and often heuristic hyperparameter tuning or model selection step.
} 
\end{abstract}


\section{Introduction}
\label{intro}

The shift from mass production to mass customization and personalization \cite{Hu.2013} requires high standards on production processes. In spite of the high variance between different products and small batch sizes of the products to be manufactured, the product quality in mass customization has to be comparable to the quality of products from established mass production processes. It is therefore essential to keep process ramp-up times low and to achieve the required product quality as directly as possible. 
This requires a profound and solid understanding of the dependencies between process parameters and quality criteria of the final product, even before the start of production (SOP). Various ways exist to gain this kind of process knowledge: for example, by carrying out experiments, setting up simulations, or exploiting available expert knowledge. In production, expert knowledge in particular plays a central role. This is because complex cause-effect relationships operate between the input-output parameters during machining, and these generally have to be set in a result-oriented manner in a short amount of time without recourse to real-time data sets. Indeed, process ramp-up is still commonly done by process experts purely based on their knowledge. Furthermore, many processes are controlled by experts during production to ensure that consistently high quality is produced.

In the course of digitalization, the acquisition of and the access to data in manufacturing have increased significantly in recent years. Sensors, extended data acquisition by the controllers themselves, and the continuous development of low-cost sensors allow for the acquisition of large amounts of data \cite{Wuest.2016}. Accordingly, more and more data-driven approaches, most notably machine learning methods, are used in manufacturing to describe the dependencies between process parameters and quality parameters \cite{Weichert.2019}. In principle, such data-driven methods are suitable for the rapid generation of quality prediction models in production, but the quality of machine learning models crucially depends on the amount and the information content of the available data. 
The data can be generated from experiments or from simulations. In general, experiments for process development or improvement are expensive and accordingly the number of experiments to be performed should be kept to a minimum. In this context, design of experiment can be used to obtain maximum information about the process behavior with as few experiments as possible \cite{Montgomery.2017}, \cite{Fedorov.2014}. Similarly, the generation of data using realistic simulation models can be expensive as well, because the models must be created, calibrated, and -- depending on the process -- high computing capacities are required to generate the data. Concluding, the data available in manufacturing before the SOP is typically rather small. 

This paper introduces a novel and general methodology to leverage expert knowledge in order to compensate such data sparsities and to arrive at prediction models with good predictive power in spite of small datasets. Specifically, the proposed methodology is dedicated to shape expert knowledge, that is, expert knowledge about the qualitative shape of the input-output relationship to be learned. Simple examples of such shape knowledge are prior monotonicity or prior convexity knowledge, for instance. Additionally, the proposed methodology directly involves process experts in capturing and in incorporating their shape knowledge into the resulting prediction model. 

In more detail, the proposed methodology proceeds as follows. In a first step, an initial, purely data-based prediction model is trained. A process expert then inspects selected, particularly informative graphs of this model and specifies in what way these graphs confirm or contradict his shape expectations. In a last step, the thus specified shape expert knowledge is incorporated into a new prediction model which strictly complies with all the imposed shape constraints. In order to compute this new model, the semi-infinite optimization approach to shape-constrained regression is taken, based on the algorithms from \cite{Schmid.2021}. In the following, this approach is referred to as the SIASCOR method for brevity. While a semi-infinite optimization approach has also been pursued in \cite{Kurnatowski.2021}, the algorithm used here is superior to the reference-grid algorithm from \cite{Kurnatowski.2021}, both from a theoretical and from a practical point of view. Additionally, the paper  \cite{Kurnatowski.2021} treats only a single kind of shape constraints, namely monotonicity constraints. 

The general methodology is applied to the exemplary process of grinding with brushes. In spite of the small set of available measurement data, the methodology proposed here leads to a high-quality prediction model for the surface roughness of the brushed workpiece.

The paper is organized as follows. Section 2 gives an overview of the related work. In Section 3, the general methodology to capture and incorporate shape expert knowledge is introduced, and its individual steps are explained in detail. Section 4 describes the application example, that is, the brushing process. Section 5 discusses the resulting prediction models applied to the brushing process and compares them to more traditional machine learning models. Section 6 concludes the paper with a summary and an outlook on future research.

\section{Some related works} 

In \cite{Weichert.2019} it is shown that machine learning models used for optimization of production processes are often trained with relatively small datasets. In this context, attempts are often made to represent complex relationships with complex models and small datasets. Also in other domains, such as process engineering \cite{Napoli.2011} or medical applications \cite{Shaikhina.2017}, small amounts of data play a role in the use of machine learning methods. Accordingly, there already exist quite some methods to train complex models with small datasets in the literature. These known approaches to sparse-data learning can be categorized as purely data-based methods on the one hand and as expert-knowledge-based methods on the other hand. In the following literature review, expert-knowledge-based approaches that typically require large -- or, at least, non-sparse --  datasets are not included. In particular, the projection- \cite{Lin.2014, Schmid.2020} and rearrangement-based \cite{Dette.2006, Chernozhukov.2009} approaches to monotonic  regression are not reviewed here.

\subsection{Purely data-based methods for sparse-data learning in manufacturing}

An important method for training machine learning models with small datasets is to generate additional, artificial data. Among these virtual-data methods the mega-trend-diffusion (MTD) technique is particularly common. It was developed by \cite{Li.2007} using flexible manufacturing system scheduling as an example. In \cite{Li.2013} virtual data is generated using a combination of MTD and a plausibility assessment mechanism. In the second step, the generated data is used to train an artificial neural network (ANN) and a support vector regression model with sample data from the manufacturing of liquid-crystal-display (LCD) panels. Using multi-layer ceramic capacitor manufacturing as an example, bootstrapping is used in \cite{Tsai.2008} to generate additional virtual data and then train an ANN. The authors of \cite{Napoli.2011} also use bootstrapping and noise injection to generate virtual data and consequently improve the prediction of an ANN. The methodology is applied to estimate the freezing point of kerosene in a topping unit in chemical engineering. In \cite{Chen.2017} virtual data is generated using particle swarm optimization to improve the prediction quality of an extreme learning machine model. 

In addition to the methods for generating virtual data and the use of simple machine learning methods such as linear regression, lasso or ridge regression \cite{Bishop.2006}, other machine learning methods from the literature can also be used in the context of small datasets. For example, the multi-model approaches in \cite{Li.2012}, \cite{Chang.2015} can be mentioned here. The multi-model approaches are used in the field of LCD panel manufacturing to improve the prediction quality. Another concrete example are the models described in  \cite{Torre.2019}, which are based on polynomial chaos expansion. These models are also suitable for learning complex relationships in spite of few data points.

\subsection{Expert-knowledge-based methods for sparse-data learning in manufacturing}

An extensive general survey about integrating prior knowledge in learning systems is given in \cite{Rueden.2021}. The integration of knowledge depends on the source and the representation of the knowledge: for example, algebraic equations or simulation results represent scientific knowledge and can be integrated into the learning algorithm or the training data, respectively. 

Apart from this general reference, the recent years brought about various papers on leveraging expert knowledge in specific manufacturing applications. Among other things, these papers are motivated by the fact that production planning becomes more and more difficult for companies due to mass customization. In order to improve the quality of production planning, \cite{Schuh.2019} show that enriching production data with domain knowledge leads to an improvement in the calculation of the transition time with regression trees.

Another broad field of research is knowledge integration via Bayesian networks. In \cite{Zhang.2020} domain knowledge is incorporated using a Bayesian network to predict the energy consumption during injection molding. In \cite{Lokrantz.2018} a machine learning framework is presented for root cause analysis of faults and quality deviations, in which knowledge is integrated via Bayesian networks. Based on synthetically generated manufacturing data, an improvement of the inferences could be shown compared to models without expert knowledge. In \cite{He.2019}  Bayesian networks are used to inject expert knowledge about the manufacturing process of a cylinder head to evaluate the functional state of manufacturing on the one hand, and to identify causes of functional defects of the final product on the other hand. Another possibility of root cause analysis using domain-specific knowledge is described by \cite{Rahm.2018}. Here, knowledge is acquired within an assistance system and combined with machine learning methods to support the diagnosis and elimination of faults occurring at packaging machines. 

In \cite{Lu.2017}, knowledge of the electrochemical micro-machining process is incorporated into the structure of a neural network. It is demonstrated that integrating knowledge achieves better prediction accuracy compared to classical neural networks. Another way to integrate knowledge about the additive manufacturing process into neural networks is based on causal graphs and proposed by \cite{Nagarajan.2019}. This approach leads to a more robust model with better generalization capabilities. In \cite{Ning.2019}, a control system for a grinding process is presented in which, among other things, a fuzzy neural network is used to control the surface roughness of the workpiece. Incorporating knowledge into models using fuzzy logic is a well-known and proven method, especially in the field of grinding \cite{Brinksmeier.2006}.

\section{A methodology to capture and incorporate shape expert knowledge}
\label{sec:methodology}

As has been pointed out in the previous section, there are expert-knowledge-free and expert-knowledge-based methods to cope with small datasets in the training of machine learning models in manufacturing. An obvious advantage of expert-knowledge-based approaches is that they typically yield models with superior predictive power, because they take into account more information than the pure data. Another clear advantage of expert-knowledge-based approaches is that their models tend to enjoy higher acceptance among process experts, because the experts are directly involved in the training of these models. 

Therefore, this paper proposes a general methodology to capture and incorporate expert knowledge into the training of a powerful prediction model for certain process output quantities of interest. Specifically, the proposed methodology is dedicated to shape expert knowledge, that is, prior knowledge about the qualitative shape of the considered output quantity $y$ as a function 
\begin{align} \label{eq:true-fct-relship}
y = y(\bm{x}) = y(x_1, \dots, x_d)
\end{align}
of relevant process input parameters $x_1, \dots, x_d$. Such shape expert knowledge can come in many forms. An expert might know, for instance, that the considered output quantity $y$ is monotonically increasing w.r.t.~$x_1$, concave w.r.t.~$x_2$, and monotonically decreasing and convex w.r.t. $x_3$.

\begin{figure}
	\centering
	\includegraphics[width=\columnwidth]{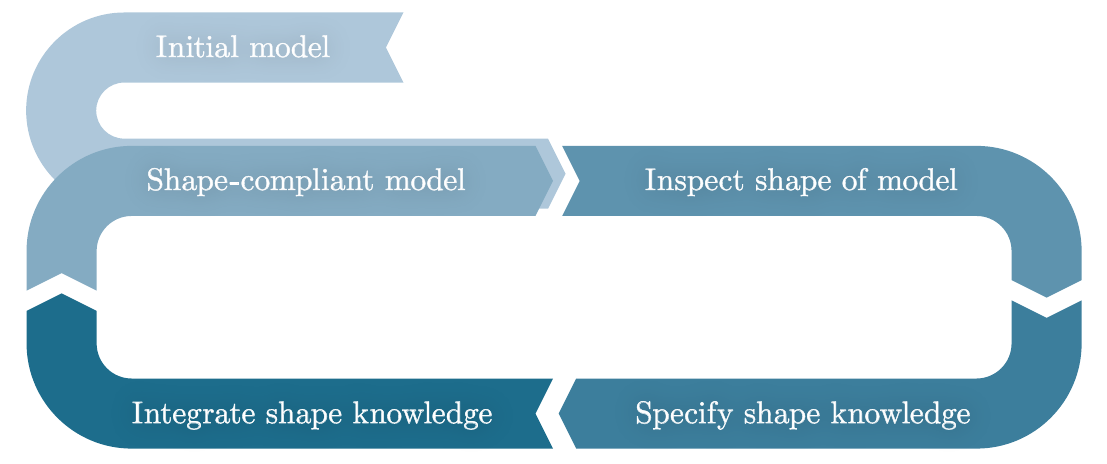}
	\caption{Schematic of the proposed methodology}
	\label{fig:methodology-schematic}
\end{figure}

In a nutshell, the proposed methodology to capture and incorporate shape expert knowledge proceeds in the following four steps. 
\begin{itemize}
	\item[1.] Training of an initial purely data-based prediction model
	\item[2.] Inspection of the initial model by a process expert
	\item[3.] Specification of shape expert knowledge by the expert
	\item[4.] Integration of the specified shape expert knowledge into the training of a new prediction model which strictly complies with the imposed shape knowledge. 
\end{itemize}
This new and shape-knowledge-compliant prediction model is computed with the help of the SIASCOR method \cite{Schmid.2021} and it is therefore referred to as the SIASCOR model. After a first run through the steps above, the shape of the SIASCOR model can still be insufficient in some respects, because the shape knowledge specified at the first run might not have been complete, yet. In this case, step two to four can be passed through again, until the expert notices no more shape knowledge violations in the final SIASCOR model. Schematically, this procedure is sketched in Figure~\ref{fig:methodology-schematic}. 

In the remainder of this section, the individual steps of the proposed methodology are explained in detail. The input parameter range on which the models are supposed to make reasonable predictions is always denoted by the symbol $X$. It is further assumed that $X$ is a rectangular set, that is,
\begin{align}
X = \{\bm{x} \in \mathbb{R}^d: a_i \le x_i \le b_i \text{ for all } i \in \{1,\dots,d\}\}
\end{align}
with lower and upper bounds $a_i$ and $b_i$ for the $i$th input parameter $x_i$. Additionally, the -- typically small -- set of measurement data available for the relationship~\eqref{eq:true-fct-relship} is always denoted by the symbol
\begin{align}
\mathcal{D} = \{(\bm{x}^j, y^j): j \in \{1, \dots, N\}\}. 
\end{align}

\subsection{Training of an initial prediction model}
\label{sec:initial-model}

In the first step of the methodology, an initial purely data-based model $\hat{y}^0$ is trained for~\eqref{eq:true-fct-relship}, using standard polynomial regression with ridge or lasso regularization \cite{Bishop.2006}. So, the initial model $\hat{y}^0$ is assumed to be a multivariate polynomial
\begin{align} \label{eq:initial-model}
\hat{y}^0(\bm{x}) = \hat{y}_{\bm{w}}^0(\bm{x}) = \bm{w}^{\top} \bm{\phi}^0(\bm{x})
\qquad (\bm{x} \in X)
\end{align}
of some degree $m^0 \in \mathbb{N}$, where $\bm{\phi}^0(\bm{x})$ is the vector consisting of all monomials $x_1^{p_1} \cdot \dotsb \cdot x_d^{p_d}$ of degree less than or equal to $m^0$ and where $\bm{w}$ is the vector of the corresponding monomial coefficients. In training, these monomial coefficients $\bm{w}$ are tuned such that $\hat{y}_{\bm{w}}^0$ optimally fits the data $\mathcal{D}$ and such that, at the same time, the ridge or lasso regularization term is not too large. In other words, one has to solve the simple unconstrained regression problem
\begin{align}
\min_{\bm{w}} \sum_{j=1}^N (\hat{y}_{\bm{w}}(\bm{x}^j) - y^j)^2 + \lambda \|\bm{w}\|_q,
\end{align}
where $\lambda \in (0,\infty)$ and $q \in \{1, 2\}$ are suitable regularization hyperparameters ($q=1$ corresponding to lasso and $q=2$ corresponding to ridge regression). 
As usual, these hyperparameters are chosen such that some cross-validation error becomes minimal.

\subsection{Inspection of the initial prediction model}
\label{sec:inspection-of-model}

In the second step of the methodology, a process expert inspects the initial model in order to get an overview of its shape. To do so, the expert has to look at $1$- or $2$-dimensional graphs of the initial model. Such $1$- and $2$-dimensional graphs are obtained by keeping all input parameters except one (two) constant to some fixed value(s) of choice and by then considering the model as a function of the one (two) remaining parameter(s). As soon as the number $d$ of inputs is larger than two, there are infinitely many of these graphs and it is notoriously difficult for humans to piece them toghether to a clear and coherent picture of the model's shape \cite{Oesterling.2016}. It is therefore crucial to provide the expert with a small selection of particularly informative graphs, namely graphs with particularly high model confidence and graphs with particularly low model confidence. 

\begin{figure}
	\centering
	\includegraphics[width=0.6\columnwidth]{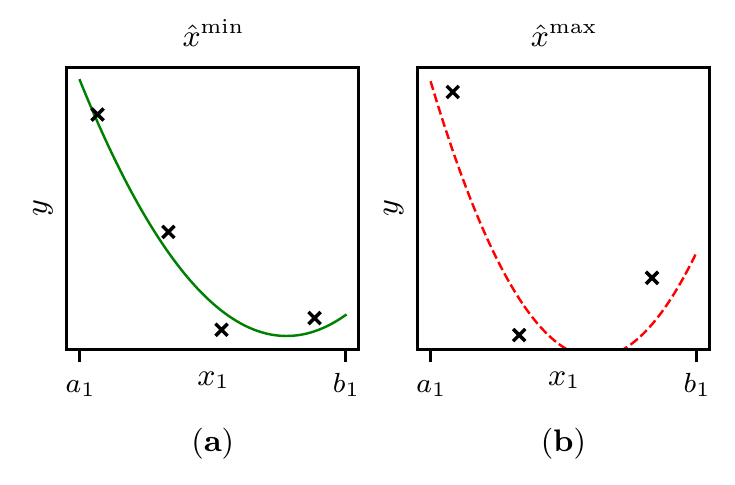} 
	\caption{Sample high- (a) and low-fidelity (b) graphs of the initial model (both in the first input dimension $x_1$)}
	\label{fig:exemplary-projections}
\end{figure}

A simple method of arriving at such high- and low-fidelity graphs is as follows. Choose those two points $\hat{\bm{x}}^{\mathrm{min}}$, $\hat{\bm{x}}^{\mathrm{max}}$ from a given grid
\begin{align} \label{eq:anchor-points-def}
\mathcal{G} = \{\hat{\bm{x}}^k: k \in \{1,\dots,K\}\}
\end{align} 
in $X$ with minimal or maximal accumulated distances from the data points, respectively. In other words, 
\begin{align} \label{eq:anchor-points-def-1}
\hat{\bm{x}}^{\mathrm{min}} := \hat{\bm{x}}^{k_{\mathrm{min}}}
\quad \text{and} \quad
\hat{\bm{x}}^{\mathrm{max}} := \hat{\bm{x}}^{k_{\mathrm{max}}},
\end{align} 
where the gridpoint indices $k_{\mathrm{min}}$ and $k_{\mathrm{max}}$ are defined by
\begin{align} \label{eq:anchor-points-def-2}
k_{\mathrm{min}} := \operatorname{argmin}_{k\in\{1,\dots,K\}} \sum_{j=1}^N \|(\bm{x}^j, y^j) - (\hat{\bm{x}}^k, \hat{y}^k)\|_2 \\
k_{\mathrm{max}} := \operatorname{argmax}_{k\in\{1,\dots,K\}} \sum_{j=1}^N \|(\bm{x}^j, y^j) - (\hat{\bm{x}}^k, \hat{y}^k)\|_2
\end{align}
with $\hat{y}^k := \hat{y}^0(\hat{\bm{x}}^k)$ being the initial model's prediction at the gridpoint $\hat{\bm{x}}^k$. Starting from the two points $\hat{\bm{x}}^{\mathrm{min}}$ and $\hat{\bm{x}}^{\mathrm{max}}$, one then traverses each input dimension range. In this manner, one obtains, for each input dimension $i$, a $1$-dimensional graph of the initial model of particularly high fidelity (namely the function $x_i \mapsto \hat{y}^0(\hat{x}_1^{\mathrm{min}}, \dots, x_i, \dots, \hat{x}_d^{\mathrm{min}})$) and a $1$-dimensional graph of particularly low fidelity (namely the univariate function $x_i \mapsto \hat{y}^0(\hat{x}_1^{\mathrm{max}}, \dots, x_i, \dots, \hat{x}_d^{\mathrm{max}})$). See Figure~\ref{fig:exemplary-projections} for exemplary high- and low-fidelity graphs as defined above. 

An alternative method of obtaining low- and high-fidelity input parameters and graphs is to use design-of-experiments techniques \cite{Fedorov.2014}, but this alternative approach is not pursued here.

After inspecting particularly informative graphs as defined above, the expert can further explore the initial model's shape by navigating through and investigating arbitrary graphs of the initial model with the help of commercial software or standard slider tools (from Python Dash or PyQt, for instance). 

\subsection{Specification of shape expert knowledge}
\label{sec:specification-of-shape-knowledge}

In the third step of the methodology, the process expert specifies his shape expert knowledge about the input-output relationship~\eqref{eq:true-fct-relship} of interest. In this process, the expert can greatly benefit from the initial model and especially from the high- and low-fidelity graphs generated in the second step. Indeed, with the help of these graphs, the expert can, on the one hand, easily detect shape behavior that contradicts his expectations and, on the other hand, identify shape behavior that already matches his expectations for the shape of~\eqref{eq:true-fct-relship}. When inspecting the graphs from Figure~\ref{fig:exemplary-projections}, for instance, the expert might notice that the initial model exceeds or deceeds physically meaningful bounds. Similarly, the expert might notice that the initial model
\begin{itemize}
\item is convex w.r.t.~$x_1$ (as he expects)
\item is not monotonically decreasing w.r.t.~$x_1$ (contrary to what he expects).
\end{itemize}
All the shape knowledge that is noticed and worked out in this manner can then be specified and expressed pictorially in the form of simple schematic graphs like the ones from Figure~\ref{fig:rebound-constraint}.

\subsection{Integration of shape expert knowledge into the training of a new prediction model} 
\label{sec:integration-of-shape-knowledge}

In the fourth step, the shape expert knowledge specified in the third step is integrated into the training of a new and shape-knowledge-compliant prediction model, using the SIASCOR method. Similarly to the initial model, the SIASCOR model $\hat{y}$ is assumed to be a multivariate polynomial
\begin{align} \label{eq:SIASCOR-model}
\hat{y}(\bm{x}) = \hat{y}_{\bm{w}}(\bm{x}) = \bm{w}^{\top} \bm{\phi}(\bm{x})
\qquad (\bm{x} \in X)
\end{align}
of some degree $m \in \mathbb{N}$ (not necessarily equal to the degree of the initial model) and $\bm{\phi}(\bm{x})$, $\bm{w}$ represent the monomials and the corresponding monomial coefficients as in~\eqref{eq:initial-model}. In contrast to the initial model training, however, the monomial coefficients $\bm{w}$ are now tuned such that $\hat{y}_{\bm{w}}$ not only optimally fits the data $\mathcal{D}$ but also strictly satisfies all the shape constraints specified in the third step. 
In other words, one has to solve the constrained regression problem
\begin{align} \label{eq:shape-constrained-regression-problem}
\min_{\bm{w}} \sum_{j=1}^N (\hat{y}_{\bm{w}}(\bm{x}^j) - y^j)^2 
\end{align}
subject to the shape constraints specified in the third step. In order to do so, the core semi-infinite optimization algorithm from \cite{Schmid.2021} is used, which covers a large variety of allowable shape constraints. 

\begin{figure}
	\centering
	\includegraphics[width=0.7\columnwidth]{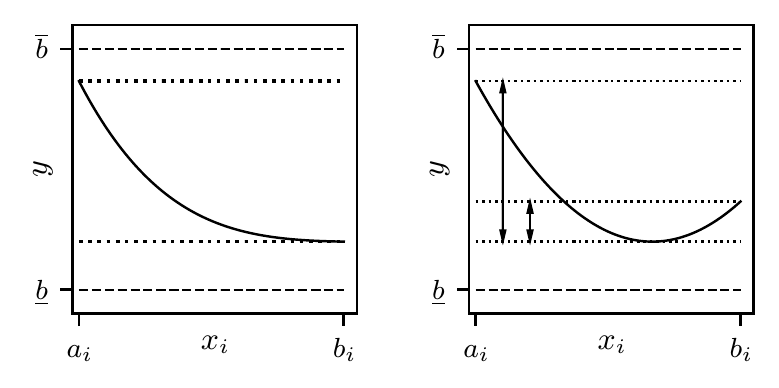}
	\caption{Sample graphs satisfying the rebound constraint with $r=1/2$}
	\label{fig:rebound-constraint}
\end{figure}

Some simple examples of shape constraints covered by the algorithm are
boundedness constraints
\begin{align} \label{eq:boundedness-constraint}
\underline{b} \le \hat{y}_{\bm{w}}(\bm{x}) \le \overline{b} \qquad (\bm{x} \in X)
\end{align}
with given lower and upper bounds $\underline{b}, \overline{b}$,
monotonic increasingness or decreasingness constraints
\begin{align} \label{eq:monotonicity-constraints}
\partial_{x_i} \hat{y}_{\bm{w}}(\bm{x}) \ge 0 \qquad (\bm{x} \in X),\\
\partial_{x_i} \hat{y}_{\bm{w}}(\bm{x}) \le 0 \qquad (\bm{x} \in X)
\end{align}
in a given input dimension $i$, 
as well as convexity or concavity constraints
\begin{align} \label{eq:convexity-constraints}
\partial_{x_i}^2 \hat{y}_{\bm{w}}(\bm{x}) \ge 0 \qquad (\bm{x} \in X),\\
\partial_{x_i}^2 \hat{y}_{\bm{w}}(\bm{x}) \le 0 \qquad (\bm{x} \in X)
\end{align}
in a specified input dimension $i$. A more complex kind of shape constraint that is also covered by the employed algorithm is the so-called rebound constraint. It constrains the amount the model can rise after a descent to be no larger than a given rebound factor $r$. In mathematically precise terms, a rebound constraint in the $i$th input dimension takes the following form:
\begin{align} \label{eq:rebound}
\hat{y}_{\bm{w}}(x_1, \dots, b_i, \dots, x_d) - \hat{y}_{\bm{w}}^* 
\le 
r \cdot (\hat{y}_{\bm{w}}(x_1, \dots, a_i, \dots, x_d) - \hat{y}_{\bm{w}}^*)
\end{align}
for all values $x_j \in [a_j,b_j]$ of the input parameters in the remaining dimensions $j \ne i$, where 
\begin{align}
\hat{y}_{\bm{w}}^* := \min_{x_i \in [a_i,b_i]} \hat{y}_{\bm{w}}(x_1, \dots, x_i, \dots, x_d)
\end{align} 
and where $r \in (0,1]$ is the prescribed rebound factor. Sample graphs of a model that satisfies this rebound constraint with $r = 1/2$ can be seen in Figure~\ref{fig:rebound-constraint}.

An important asset of the approach to shape-con-strained regression taken here is that the core algorithm can handle arbitrary combinations of the kinds of shape constraints mentioned above, in an efficient manner. Also, the core algorithm is entirely implemented in Python which makes it particularly easy to use and interface.
Another asset of the proposed approach is that the considered shape-constrained regression problem~\eqref{eq:shape-constrained-regression-problem} features no hyperparameter except for the polynomial degree $m$. Consequently, no tuning of hard-to-interpret hyperparameters is necessary. Concerning other, more theoretical, merits of the employed semi-infinite optimization algorithm, the reader is referred to \cite{Schmid.2021}.

\section{Application example} 
\label{sec:application-example}

\subsection{The brushing process}
\label{sec:brushing-process}

The brushing process is a metal-cutting process used for the grinding of metallic surfaces with the help of brushes. Its main applications are the deburring of precision components \cite{Gillespie.1979}, the structuring of decorative surfaces of glass \cite{Novotny.2017}, and the functional surface preparation of metals for subsequent process steps of joining \cite{Teicher.2018}. Common to all these applications is that the brushing process functions as a finishing process for components with a high inherent added value. Additionally, brushing processes have established themselves in certain highly automated mass production processes \cite{Kim.2012}.

\begin{figure}
	\centering
	\includegraphics[width=0.8\columnwidth]{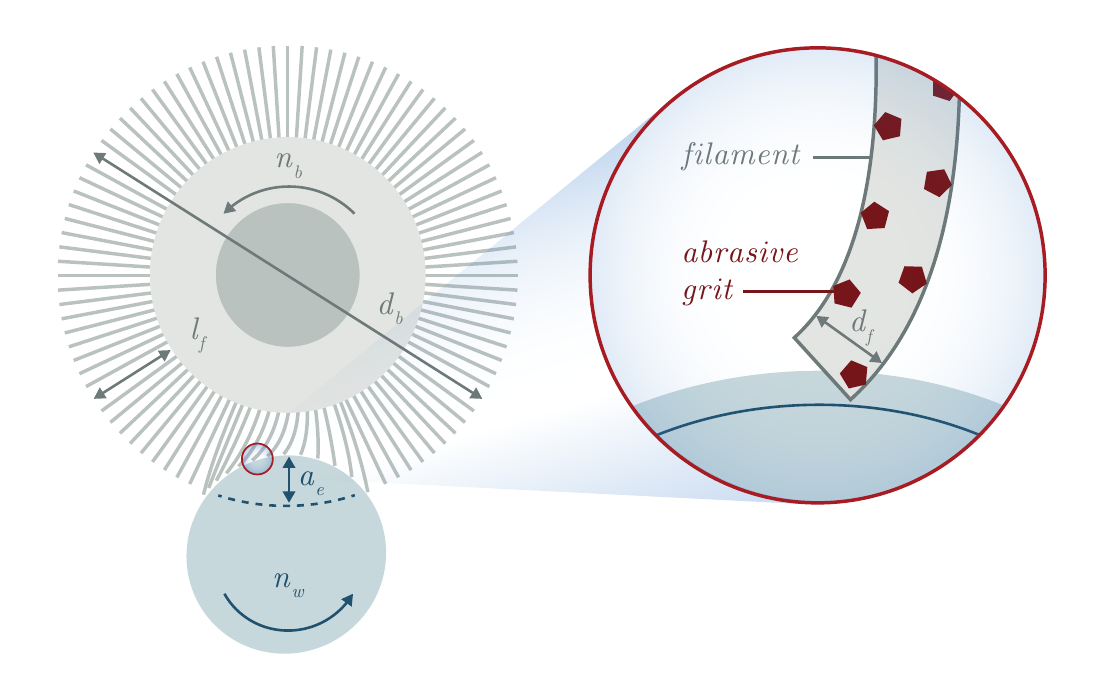}
	\caption{Schematic of the brushing process}
	\label{fig:application-example}
\end{figure}

While the focus of \cite{DIN8589} is still on steel wires as brushing filaments, in recent years filaments made of plastic with interstratified abrasive grits have become much more important. Such filaments act only as carrier elements of the machining substrate and, accordingly, the corresponding brushing process can be classified as a process with a geometrically undefined cutting edge. In view of their increased relevance, only brushing filaments with interstratified abrasive grits are considered here. See Figure~\ref{fig:application-example} for a schematic representation of the considered brushing processes. 

Apart from the material parameters of the workpiece, the machining process is influenced, on the one hand, by technological parameters of the process and, on the other hand, by a multitude of material parameters of the brush. Important technological parameters are the numbers of revolutions $n_b$ and $n_w$ of the brush and of the workpiece, the cutting depth $a_e$, and the cutting time $t_c$. The brush parameters relate to the individual filaments (length $l_f$, diameter $d_f$, modulus of elasticity, and other technical properties), their arrangement (axial, radial), and their coupling to the base body (cast, plugged). The cutting substrate as an abrasive grain is characterized, among other things, by the grain material, the grain concentration and the grain diameter $dia$. In addition, the shape of the brush is determined by its width and its diameter $d_b$. 

In view of this large variety of technological and material parameters, it is a challenging task to choose the tool and the tool settings such that a prescribed target value for the roughness of the brushed workpiece is reached quickly but also robustly. It is therefore important to have good prediction models for the surface roughness of the brushed workpiece.

In principle, such prediction models can be obtained from a comprehensive simulation of the brushing process \cite{Wahab.2007}, \cite{Novotny.2017}. Such simulation-based models are expensive and complex because -- in addition to the many process parameters mentioned above -- the dynamic behavior of the tool has to be broken down to the filaments and microscopically to the individual grain in engagement. In particular, the dynamically changing tool diameter \cite{Matuszak.2015} has to be taken into account. In addition to the challenging modeling procedure, the resulting models are typically expensive to evaluate. Currently, these factors still limit the applicability of simulation-based models in real-world process design and process control. And therefore it is important to build good alternative prediction models in brushing, for example, by using machine learning.

\subsection{Input parameters, output parameter, and dataset} 
\label{sec:input-output-parameters-and-data-set}

In this paper, such an alternative, machine learning model is built. Specifically, the modeled output quantity is the arithmetic-mean surface roughness of the brushed workpiece,
\begin{align}
y := R_a.
\end{align}
It is modeled as a function of $5$ particularly important process parameters $\bm{x} = (x_1, \dots, x_5)$ of the brushing process, namely 
\begin{align}
\bm{x} = (x_1, \dots, x_5) := (dia, \, t_c, \, n_b, \, n_w, \, a_e).
\end{align}
The dataset used for the training of the prediction model consists of $N = 125$ measurement points. Table \ref{tab:1} shows the ranges of the process and quality parameters covered by the measurement data.

\begin{table}
\centering
	\caption{The ranges of the process input and output parameters}
	\label{tab:1}
	\begin{tabular}{c|p{8.5cm}|c|c}
		\hline\noalign{\smallskip}
		symbol & process parameter & value range & unit  \\
		\noalign{\smallskip}\hline\noalign{\smallskip}
		$ dia $ 	& diameter of the abrasive grits, expressed in terms of the mesh size & $ 400 - 800 $ 	& $ - $ \\
		$ t_c $ 	& cutting time, that is, the time the brush is engaged, including contact with the workpiece & $ 15 - 480 $ 	& $ \mathrm{s} $ \\
		$ n_b $ 	& number of revolutions of the brush 	& $ 1000 - 2500 $ 	& $ \mathrm{min}^{-1} $ \\
		$ n_w $ 	& number of revolutions of the workpiece 	& $ 100 - 1000 $ 	& $ \mathrm{min}^{-1} $ \\
		$ a_e $ 	& cutting depth 	& $ 0.25 - 1.0 $ 	& $\mathrm{mm}$ \\
		$ R_a $ 	& arithmetic-mean roughness 	& $ 0.14 - 0.30 $ 	& $ \mu \mathrm{m} $ \\
		\noalign{\smallskip}\hline
	\end{tabular}
\end{table}

\section{Results and discussion}
\label{sec:results}

In this section, SIASCOR is applied to the brushing process example. In particular, shape expert knowledge is integrated according to the methodology described in Section~\ref{sec:methodology}. Aside from SIASCOR,  a purely data-driven Gaussian process regression (GPR) was conducted for the brushing example. In the end, the two regression models are compared and their advantages and shortcomings are discussed.

\subsection{Initial model}

As a first step, an initial purely data-based model was trained to visually assist the process expert in specifying shape knowledge for the SIASCOR model. A polynomial model~\eqref{eq:initial-model} with the relatively small degree $m^0=3$ was used to prevent an overfit to the small dataset. The parameters of the model were computed via lasso regression with a learning rate $\lambda$ selected by means of cross-validation using scikit-learn \cite{Pedregosa.2011}. Additionally, prior to training the input variables were transformed with the standard transformation \cite{Kuhn.2013} 
$x_i' := \sqrt{x_i}$ for all $i=1, \dots, 5$ and then scaled to the unit hypercube. The standard transformation with the square root function lead to a better generalization performance.

\subsection{Capturing shape exptert knowledge}
\label{sec:results_discussion:capturing_expert_knowledge}

\begin{figure}
\centering
\includegraphics[width=0.7\columnwidth]{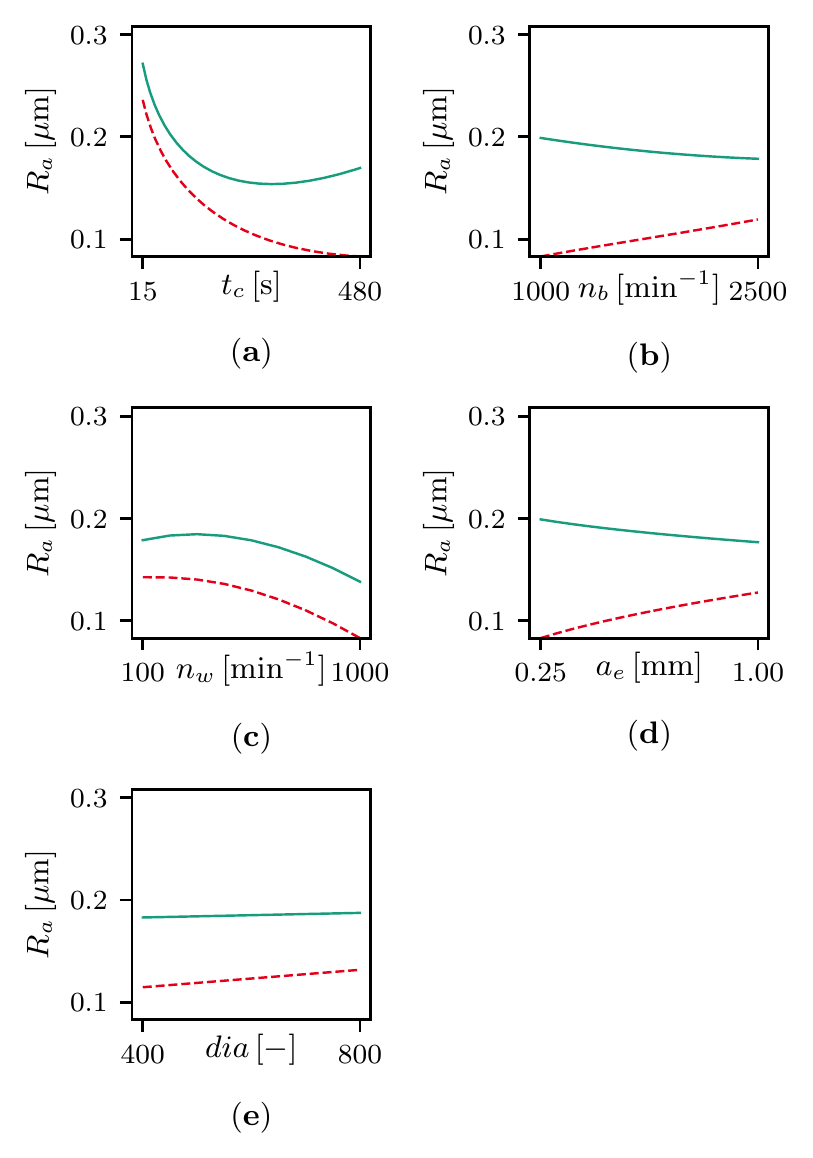}
\caption{Comparison of the high- (solid green) and low-fidelity (dashed red) graphs of the initial model (anchored in the point $\hat{\bm{x}}^{\mathrm{min}}$ or $\hat{\bm{x}}^{\mathrm{max}}$, respectively, from Table~\ref{tab:2}).}
\label{fig:projections-initial-model}
\end{figure}

\begin{figure}
\centering
\includegraphics[width=0.6\columnwidth]{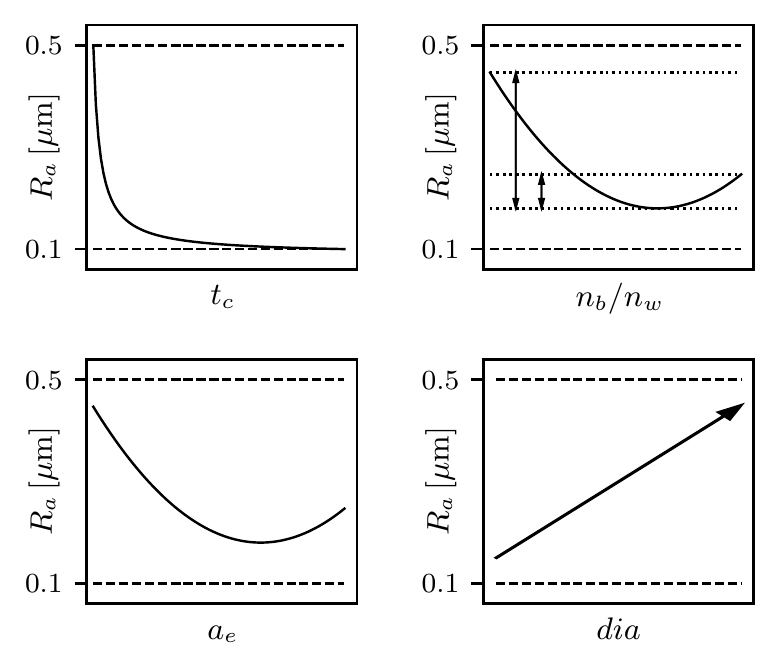}
\caption{Shape constraints specified by the process expert}
\label{fig:shape-constraints-expert}
\end{figure}

As a second step, for the inspection of the initial model, two points $\hat{\bm{x}}^{\mathrm{min}}$, $\hat{\bm{x}}	^{\mathrm{max}}$ of particularly high fidelity and of particularly low fidelity were computed according to~\eqref{eq:anchor-points-def}-\eqref{eq:anchor-points-def-2} (Table~\ref{tab:2}). The corresponding $1$-dimensional graphs of the initial model (anchored in these two points) are visualized in Figure~\ref{fig:projections-initial-model}. When inspecting and analyzing the shape of these graphs, the process expert detected several physical inconsistencies. For example, some of the initial model's predictions for $R_a$ are significantly lower than the surface roughness that is technologically achievable with the brushing process. Another example is the violation of convexity along the $n_w$ direction. With these observations in mind, the expert specified shape constraints for the SIASCOR model in the form of the schematic graphs from Figure~\ref{fig:shape-constraints-expert}. Specifically, the expert imposed the boundedness constraint $0.1 \leq R_a \leq 0.5$ upon the surface roughness. Along the $t_c$ direction, the expert required monotonic decreasingness and convexity. In the direction of $n_b$ and $n_w$, the model was required to be convex and to satisfy the rebound constraint~\eqref{eq:rebound} with $r = 1/2$. And finally, the model was constrained to be convex w.r.t.~$a_e$ and monotonically increasing w.r.t.~$dia$.

\begin{table}
\centering
	\caption{Anchor points $\hat{\bm{x}}^{\mathrm{min}}$ and $\hat{\bm{x}}^{\mathrm{max}}$ for the high- and low-fidelity graphs}
	\label{tab:2} 
	\begin{tabular}{c|c|c|c|c|c}
		\hline\noalign{\smallskip}
		point & $ dia $ & $ t_c \, \mathrm{[s]} $ & $ n_b \, \mathrm{[min^{-1}]} $ & $ n_w \, \mathrm{[min^{-1}]} $ & $ a_e \, \mathrm{[mm]} $ \\
		\noalign{\smallskip}\hline\noalign{\smallskip}
		$ \hat{\bm{x}}^{\mathrm{min}} $	&400  &106  &1964  &438    &0.84 \\
		$ \hat{\bm{x}}^{\mathrm{max}} $ 	&800  &480  &1000  &1000  &0.25 \\
		\noalign{\smallskip}\hline
	\end{tabular}
\end{table}

\subsection{SIASCOR model}

With the aforementioned shape constraints and the data described in Section~\ref{sec:input-output-parameters-and-data-set}, the SIASCOR model was trained as explained in Section~\ref{sec:integration-of-shape-knowledge}. For the degree of the polynomial model, $m=4$ was found to be the best fit. Moreover, the variables were transformed with the root function $x_i' = \sqrt{x_i}$ for all $i=1, \dots, 5$ and then scaled to the unit hypercube. Table~\ref{tab:performance-measures-SIASCOR-and-GPR} lists various performance indices and Figure~\ref{fig:prediction-SIASCOR} shows two plots of the final SIASCOR model.

\subsection{GPR model}

In addition to the SIASCOR model, a GPR model was trained for the sake of comparison since GPR with an appropriately chosen kernel is well-suited for small datasets. As a kernel, the sum of an anisotropic Mat\'{e}rn kernel with $\nu = 3/2$ and a white-noise kernel was chosen: 
\begin{align}
k(\bm{x},\bm{x}') 
&= k_{\bm{l}}(\bm{x},\bm{x}') + k_{n}(\bm{x},\bm{x}') \\
&= ( 1 + \sqrt{3} \|\bm{x}-\bm{x}'\|_{2,\bm{l}} ) \exp ( -\sqrt{3} \|\bm{x}-\bm{x}'\|_{2,\bm{l}} ) 
+ n \cdot \delta_{\bm{x},\bm{x}'} \notag
\end{align}
where $\| \bm{z} \|_{2,\bm{l}} := \|(z_1/l_1, \dots, z_d/l_d)\|_2$ denotes the aniso-tropic norm of the $d$-component vector $\bm{z}$ and where $\delta_{\bm{x},\bm{x}'}$ is $1$ if $\bm{x}=\bm{x}'$ and $0$ otherwise. 
As usual, to optimize the hyperparameters $ \bm{l} $ and $ n $, the marginal likelihood was maximized according to \cite{Williams.2006}, using the Python package scikit-learn \cite{Pedregosa.2011}. Due to the anisotropy of the Mat\'{e}rn kernel, for each input dimension $i$, a separate hyperparameter $l_i $ is calculated. As for the SIASCOR model, the input variables were transformed with the root function $x_i' = \sqrt{x_i}$ for all $i=1, \dots, 5$ and then scaled to the unit hypercube. Table~\ref{tab:performance-measures-SIASCOR-and-GPR} reports the pertinent performance indices and Figure~\ref{fig:prediction-GPR} shows two plots of the final GPR model.

\subsection{Comparison of SIASCOR and GPR}

\begin{figure}[htbp]%
\centering
\begin{subfigure}[b]{0.49\textwidth}
\centering
\includegraphics[width=\columnwidth]{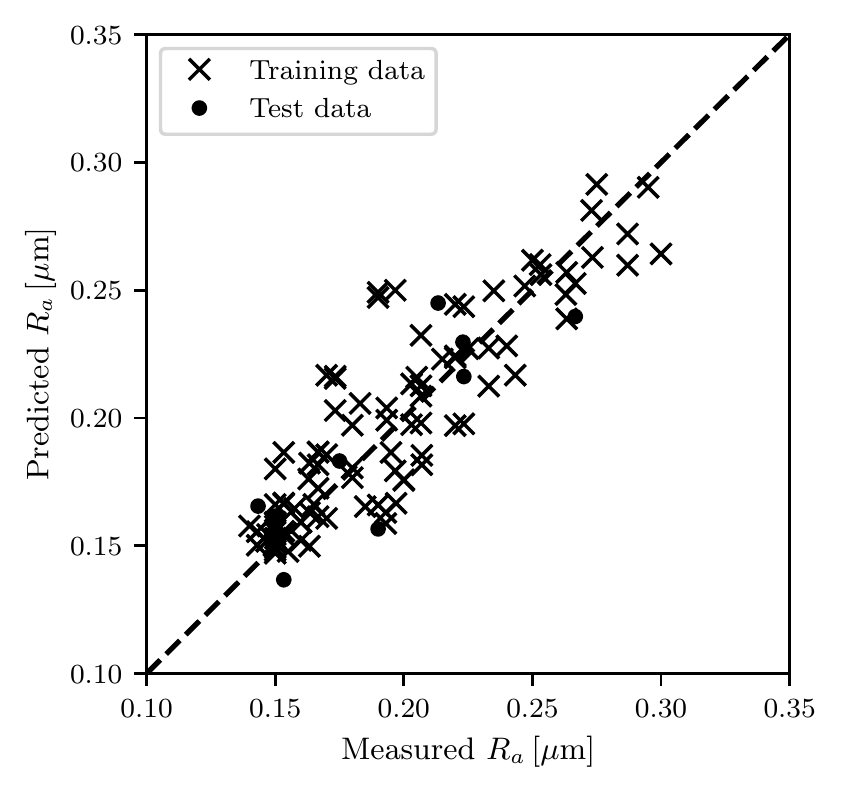}
\end{subfigure}
\hfill
\begin{subfigure}[b]{0.49\textwidth}
\centering
\includegraphics[width=\columnwidth]{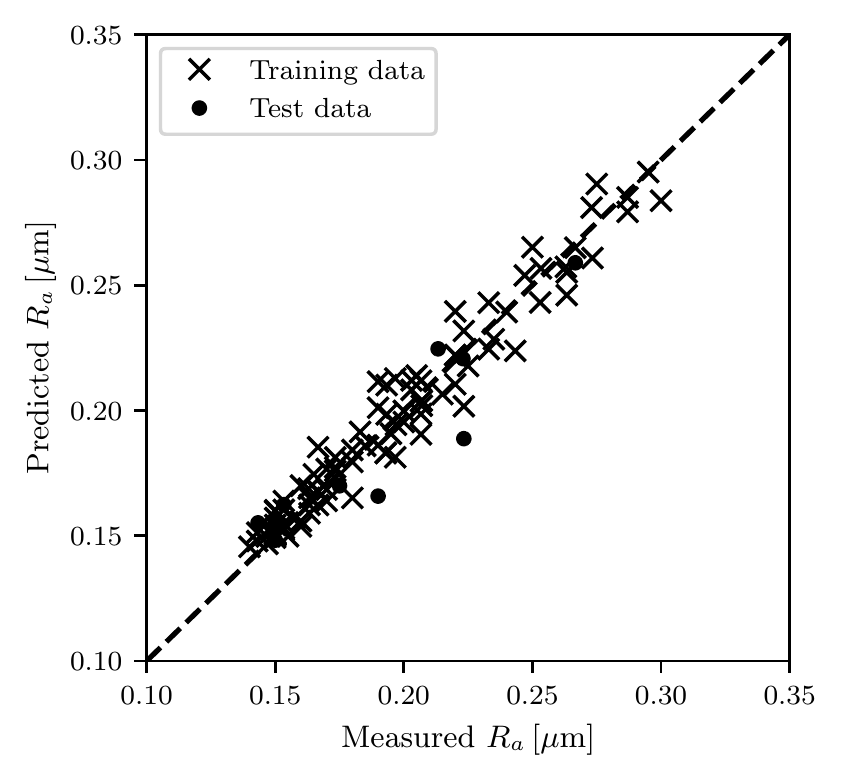}
\end{subfigure}
\caption{Comparison of measured vs. predicted values (a) for the SIASCOR model and (b) for the GPR model}
\label{fig:predictions-vs-measurements}
\end{figure}

Table~\ref{tab:performance-measures-SIASCOR-and-GPR} compares the predictive power of the initial lasso, the SIASCOR and the GPR model on test data obtained by $10$-fold cross-validation. It can be seen that the lasso and the SIASCOR models have similar averaged prediction errors and a similar averaged coefficient of determination on the test data, while the purely data-based GPR model features slightly better prediction errors. 
This can also be seen from Figure~\ref{fig:predictions-vs-measurements}. 

\begin{table}
\centering
	\caption{Various averaged performance measures for the SIASCOR and the GPR models based on $10$-fold cross-validation: root-mean-square error (RMSE), mean-absolute error (MAE), coefficient of determination ($R^2$)}
	\label{tab:performance-measures-SIASCOR-and-GPR}
	\begin{tabular}{c|c|c|c}
		\hline\noalign{\smallskip}
		model & RMSE [$\mu \mathrm{m}$] & MAE [$\mu \mathrm{m}$] & $R^2$ [--]\\
		\noalign{\smallskip}\hline\noalign{\smallskip}
		Lasso & 0.0272 & 0.0205 & 0.8353 \\
		SIASCOR & 0.0260 & 0.0193 & 0.8284 \\
		GPR & 0.0174 & 0.0142 & 0.7410 \\
		\noalign{\smallskip}\hline
	\end{tabular} 
\end{table}

\vfill

%

\begin{figure}[htbp]%
\centering
\begin{subfigure}[b]{0.49\textwidth}
\centering
\includegraphics[width=\columnwidth]{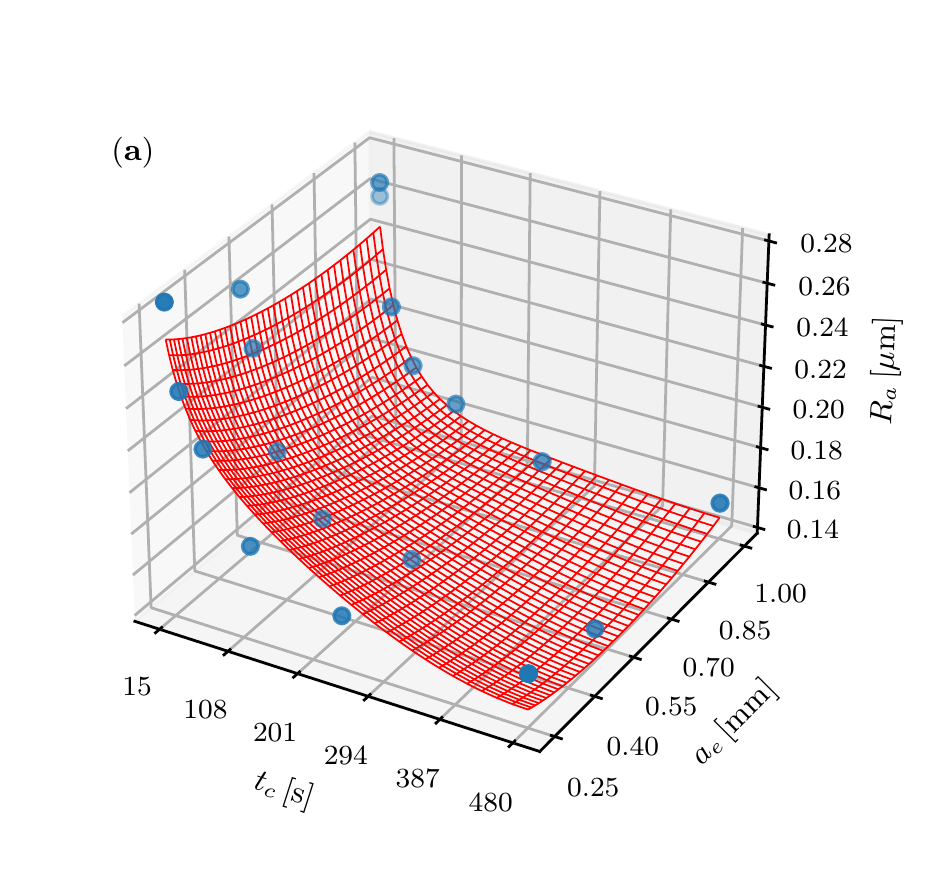}
\end{subfigure}
\hfill
\begin{subfigure}[b]{0.49\textwidth}
\centering
\includegraphics[width=\columnwidth]{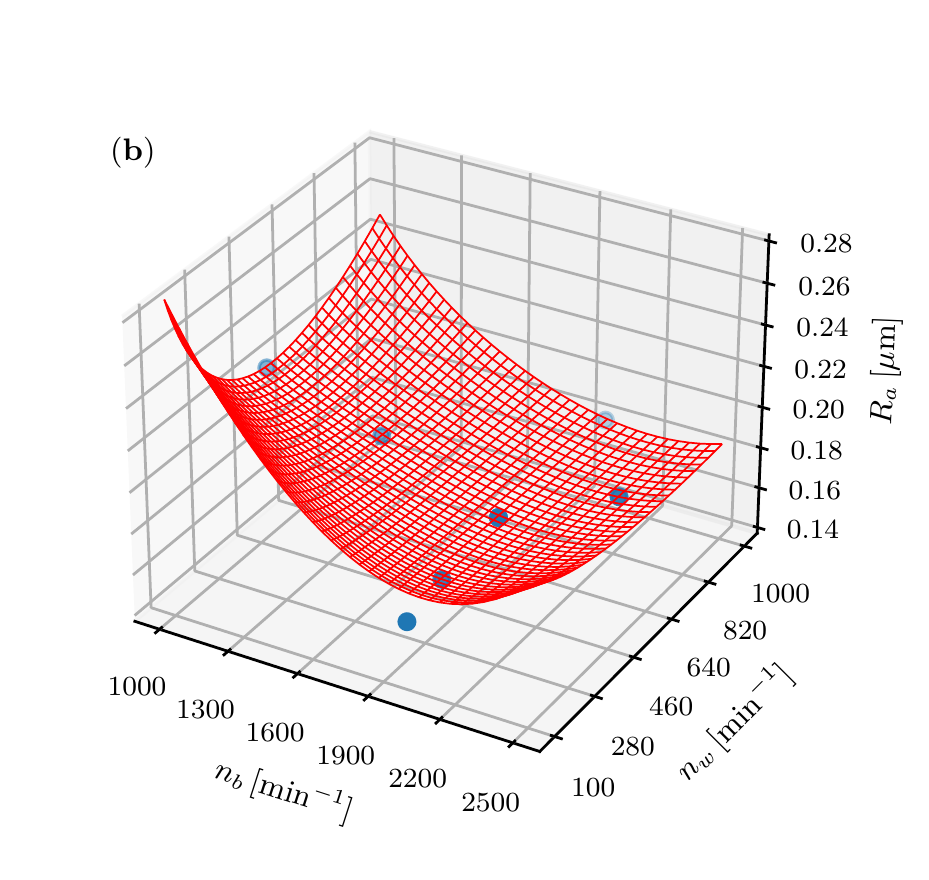}
\end{subfigure}
\caption{Sample $2$-dimensional graphs of the SIASCOR model (data points in blue). In (a), $(dia, n_b, n_w)$ are fixed to $(400, 2000, 500)$ and in (b), $(dia, t_c, a_e)$ are fixed to $(400, 60, 1)$}
\label{fig:prediction-SIASCOR}
\end{figure}

\begin{figure}[htbp]%
\centering
\begin{subfigure}[b]{0.49\textwidth}
\centering
\includegraphics[width=\columnwidth]{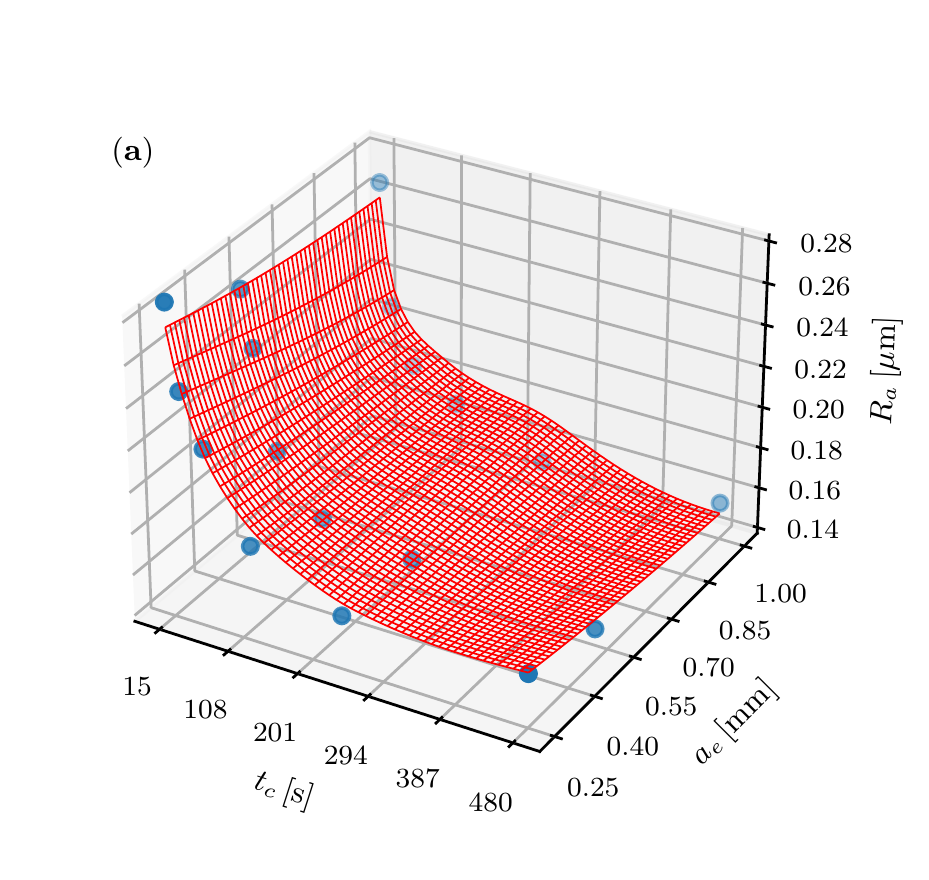}
\end{subfigure}
\hfill
\begin{subfigure}[b]{0.49\textwidth}
\centering
\includegraphics[width=\columnwidth]{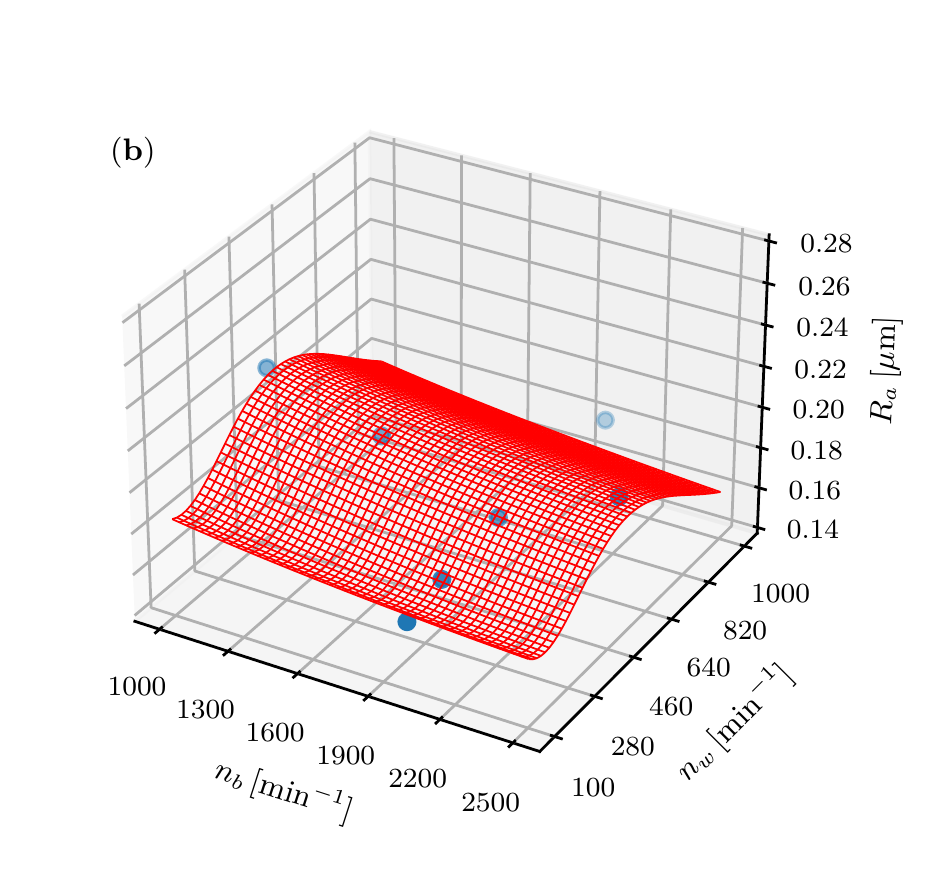}
\end{subfigure}
\caption{Sample $2$-dimensional graphs of the GPR model (data points in blue). In (a), $(dia, n_b, n_w)$ are fixed to $(400, 2000, 500)$ and in (b), $(dia, t_c, a_e)$ are fixed to $(400, 60, 1)$}
\label{fig:prediction-GPR}
\end{figure}

Figures~\ref{fig:prediction-SIASCOR} and~\ref{fig:prediction-GPR} juxtapose two plots of the SIASCOR and the GPR model, respectively. As can be seen, in contrast to the SIASCOR model, the GPR model is starkly non-convex w.r.t.~$n_w$. In other words, the GPR model is at odds with physical shape expert knowledge, while the SIASCOR model is not. As has been explained in Section~\ref{sec:methodology}, the reason is that SIASCOR explicitly incorporates all the shape knowledge provided by the process expert, while the GPR model relies on the scarce data alone. 

Another downside of the GPR approach is that the resulting models are typically quite sensitive w.r.t.~the selected kernel class and that the selection of this kernel class is typically not very systematic but rather based on heuristic rules of thumb. Accordingly, the model selection in GPR is typically quite time-consuming and cumbersome. In the SIASCOR method, by contrast, model selection is simple because the SIASCOR models have only one hyperparameter, namely the polynomial degree $m$. Also, the interpretation of the shape constraints needed for the SIASCOR method is straightforward and, in any case, much clearer than the interpretation and selection of different GPR kernel classes. 

As a matter of fact, the solution of the SIASCOR training problem~\eqref{eq:SIASCOR-model} with the algorithm from \cite{Schmid.2021} takes a bit more computational time than the hyperparemter optimization in GPR because semi-infinite optimization problems have a more complex (bi-level) structure than the (unconstrained) marginal likelihood maximization problems used in GPR. Indeed, in the $5$-dimensional brushing example considered here, the training of the SIASCOR model typically took around $30$ minutes calculated with a standard office computer. Yet, this is negligible in view of the aforementioned clear advantages of SIASCOR over GPR in terms of shape-knowledge compliance, model selection, and interpretability. 

\vfill

\vfill



\section{Conclusion and future work}

In order to achieve target product qualities quickly and consistently in manufacturing, reliable prediction models for the quality of process outcomes as a function of selected process parameters are essential. Since the datasets available in manufacturing -- and especially before SOP -- are typically small, the construction of data-driven prediction models is a challenging task. The present paper addresses this challenge by systematically leveraging expert knowledge. Specifically, this paper introduces a general methodology to capture and incorporate shape expert knowledge into machine learning models for quality prediction in  manufacturing. 

It consists of four steps:  1. training of an initial purely data-based prediction model, 2. inspection of the initial model by a process expert, 3. specification of shape expert knowledge, and 4. integration of the specified shape expert knowledge into the training of a new prediction model that complies with the shape knowledge. In the second step, the expert may find inconsistencies in the shape of the initial model contradicting the expected shape behavior. Therefore, in the third step, the expert can constrain the shape of the model to behave as expected. It is possible to define and combine as many shape constraints as desired. In the fourth step, the specified shape constraints are passed to the SIASCOR algorithm. 

The resulting SIASCOR model is mathematically guaranteed to satisfy all the shape constraints imposed by the expert. Conventional purely data-based models, by contrast, do not come with such a guarantee but, on the contrary, often exhibit an unphysical shape behavior in the sparse-data case considered here. 
Additionally, the direct involvement of process experts in the training of the SIASCOR model increases the acceptance of and the confidence in this model. Another asset of the SIASCOR method is that, in contrast to many conventional machine learning methods, it does not involve a time-consuming and unsystematic hyperparameter tuning or model selection step. 

The proposed general methodology was applied to an exemplary brushing process in order to obtain a prediction model for the arithmetic-mean surface roughness of the brushed workpiece as a function of five process parameters. The dataset available in this application consisted of only $125$ measurement points. After inspecting the initial lasso model based solely on these data, the expert defined shape constraints in all five input parameter dimensions. The SIASCOR model trained with these shape constraints was compared to a purely data-based GPR model. As opposed to the SIASCOR model, the GPR model contradicts the physical shape knowledge about the surface roughness in various ways. 
Also, the selection of an appropriate GPR kernel class is rather heuristic and time-consuming. In any case, the interpretation of the GPR kernel class is certainly less clear than the interpretation of the shape constraints used in the SIASCOR method. 

A possible topic of future research is to develop a more sophisticated definition of high- and low-fidelity graphs, using techniques from the optimal design of experiments. Another topic of future research is the further improvement of the SIASCOR algorithm's runtimes. In addition, a methodology will be developed for assessing the model and for uncovering possible conflicts between the imposed shape constraints and the data. Such conflicts might arise especially as soon as more data is available after SOP, and the model can then be retrained. And finally, a graphical user interface will be implemented allowing the domain experts to apply the proposed methodology completely independently of external support from data scientists or mathematicians. In particular, this user interface will no longer require a manual translation of the shape knowledge specified pictorially by the expert into mathematical constraints in the form expected by the SIASCOR algorithm.

\section*{Acknowledgments}
We gratefully acknowledge the funding provided by the Fraunhofer Society as part of the lighthouse project \enquote{Machine Learning for Production} (ML4P). In addition, we would like to thank Holger P\"atzold (Schaeffler Technologies AG \& Co KG, Herzogenaurach) for the valuable discussions regarding the brushing process, Markus Renner (Carl Hilzinger-Thum GmbH \& Co. KG, Tuttlingen) for providing the brushing tools and Konstantin Kusch (Fraunhofer IWU, Chemnitz) for data acquisition and analysis. We would also like to thank Michael Bortz, Jan Schwientek, and Philipp Seufert (Fraunhofer ITWM, Kaiserslautern) for inspiring mathematical discussions.

\bibliographystyle{abbrv}
{\small 
\bibliography{bibliography.bib}
}

\end{document}